# Multimodal Deep Learning for ATCO Command Lifecycle Modeling and Workload Prediction

Kaizhen Tan*
*School of Economics and Management*
*Tongji University*
Shanghai, China
tkz@tongji.edu.cn
*Corresponding author

*Abstract*—Air traffic controllers (ATCOs) issue high-intensity voice commands in dense airspace, where accurate workload modeling is critical for safety and efficiency. This paper proposes a multimodal deep learning framework that integrates structured data, trajectory sequences, and image features to estimate two key parameters in the ATCO command lifecycle: the time offset between a command and the resulting aircraft maneuver, and the command duration. A high-quality dataset was constructed, with maneuver points detected using sliding window and histogram-based methods. A CNN-Transformer ensemble model was developed for accurate, generalizable, and interpretable predictions. By linking trajectories to voice commands, this work offers the first model of its kind to support intelligent command generation and provides practical value for workload assessment, staffing, and scheduling.

## I. Introduction

### A. Background

As global air traffic demand increases, airspace operations have become more complex and congested, presenting major challenges for air traffic control (ATC) systems. Although surveillance and communication technologies have improved, ATC performance still largely depends on human operators, particularly air traffic controllers (ATCOs), who monitor flights, assess conditions, and issue maneuver instructions to ensure safe and efficient operations. With rising traffic, ATCOs face growing cognitive demands and heavier workloads. This human bottleneck has become a key constraint on ATC efficiency and safety, emphasizing the importance of quantifying task intensity and evaluating workload to support fatigue management, staff scheduling, and the development of intelligent ATC solutions.

Early studies on ATCO workload modeling primarily focused on statistical methods and subjective assessments such as NASA Task Load Index (NASA-TLX) [1]. The Dynamic Density concept by Laudeman et al. (1998) [2] estimated workload via linear combinations of traffic metrics. Tobaruela et al. (2014) [3] proposed using command frequency and type as workload proxies. While interpretable, these models are limited by handcrafted features and lack flexibility in dynamic environments. Machine learning approaches have improved adaptability by incorporating more features and learning nonlinear relationships. Gianazza et al. (2017) [4] inferred workload from traffic complexity using models like neural networks and Gradient Boosted Trees. Other studies [5] used operational data to train Random Forest and XGBoost models for workload-related predictions. However, most rely on engineered features and overlook the spatial–temporal structure of ATC operations.

Deep learning has recently shown promise in ATCO workload modeling due to its capacity to learn representations from high-dimensional, multimodal data. CNNs, adept at capturing spatial patterns, have been used to model airspace complexity via grid-based inputs. For example, Xie et al. (2021) [6] converted real-time traffic data into multichannel scene images, enabling CNN-based sector complexity prediction without manual features. Graph-based methods, such as those by Pang et al. (2023) [7], modeled aircraft interactions as dynamic graphs, leveraging GCNs to infer workload from evolving topologies. To capture temporal dynamics, RNNs have been applied. Shyr et al. (2024) [8] used LSTMs to forecast workload trends via time-series indicators. Transformers have also gained traction due to their attention mechanisms; Yang et al. (2023) [9] combined CNN and Transformer encoders to extract spatial-temporal features from Mel-spectrograms, achieving 97.48% accuracy in cognitive load estimation. In parallel, multimodal data fusion has become an increasingly important direction, with several studies integrating radar trajectories, voice communication, and controller action data into unified learning frameworks.

Despite growing interest in modeling ATCO workload, many machine learning approaches neglect key contextual factors such as weather, airspace structure, and aircraft interactions, limiting realism and generalizability. Traditional methods based on surveys or psychological models also fail to capture the dynamic, task-driven nature of ATCO operations. In practice, the issuance of control commands follows a structured temporal sequence—from situational perception and command delivery to aircraft execution and potential follow-up—collectively referred to here as the ATCO command lifecycle. To address this, the proposed study models two temporal variables within this lifecycle: *Time Offset* (delay between command issuance and aircraft response) and *Duration* (length of the spoken command). These variables help reconstruct the controller's timeline and enable real-time, interpretable workload prediction through closed-loop behavioral modeling.

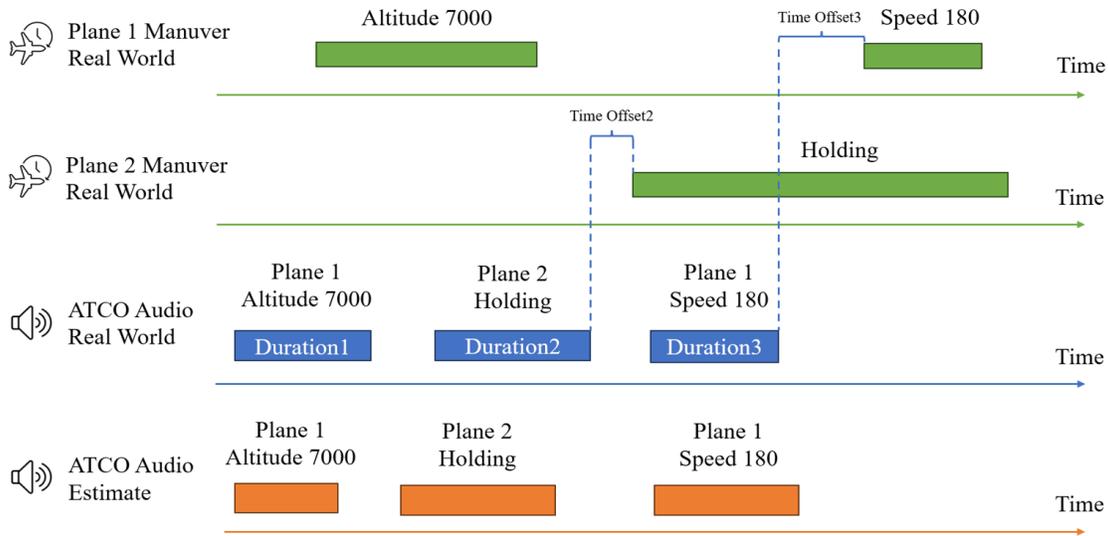

Fig. 1. Comparison of actual and estimated ATCO command lifecycles.

## B. Problem Definition

This study aims to predict the ATCO command lifecycle in terminal airspace, focusing on estimating workload through temporal behavior analysis. The lifecycle spans from the issuance of a spoken command to the execution of the corresponding aircraft maneuver, and is defined by two key variables: Time Offset and Duration, as illustrated in Fig. 1.

*1) Time Offset:* Time Offset is the delay between command issuance and maneuver initiation, capturing pilot response time and system latency. As shown in Fig. 1, many maneuvers have noticeable delays (Time Offset2, Time Offset3), while others, like pre-planned altitude changes, may begin with minimal delay or even before command completion. These variations arise from factors such as scheduling or pilot habits. Accurate Time Offset prediction is essential for reconstructing controller timelines and assessing real-time demand. This study addresses it via a deep learning model to identify command completion points on the audio timeline.

*2) Duration:* Duration refers to how long an ATCO command remains audible, indirectly reflecting its complexity and information content. As shown in Fig. 1 (Duration1, 2, 3), longer utterances often imply multi-task commands involving speed, altitude, or heading changes. In the framework, the predicted command end time (from the Time Offset model) minus Duration yields the precise issuance time, enabling reverse mapping from behavior to voice. By jointly modeling Time Offset and Duration, the proposed approach captures key temporal features of the ATCO command lifecycle.

## C. Significance and Contributions

This study primarily addresses the prediction of ATCO spoken commands and represents, to the best of current knowledge, the first attempt to infer controller command timelines directly from aircraft trajectories and airspace context. Central to this framework is the joint prediction of two key temporal variables: Time Offset (delay between command issuance and aircraft execution) and Duration (length of the spoken command). These variables enable reconstruction of the controller's operational timeline, offering a structured, data-driven representation of task intensity.

A new paradigm for ATC task modeling is proposed, allowing for dynamic estimation of controller workload in terminal maneuvering areas. By predicting when commands are issued and how long they persist, the model simulates ATCO behavior under varying traffic conditions. As illustrated in Fig. 2(a), overlapping command durations highlight periods of concurrent demand, critical for staffing and task allocation. Fig. 2(b) shows that the cumulative duration of commands provides a direct and interpretable workload metric. Additional indicators, such as command frequency and interval, further support cognitive demand assessment and fatigue monitoring.

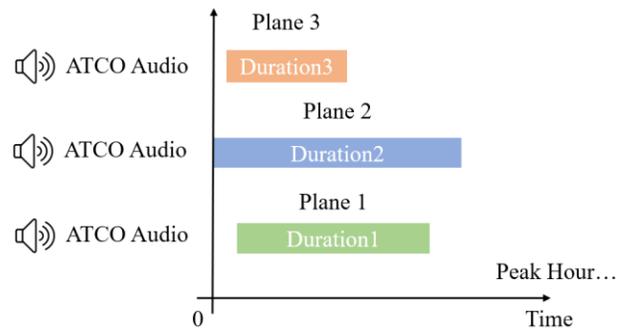

Fig. 2(a). Command overlap in peak hour.

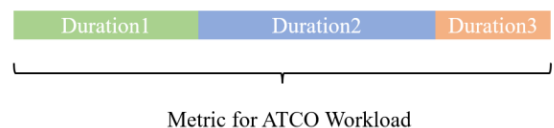

Fig. 2(b). Aggregated command duration as workload metric.

The key contributions of this study are as follows:

*1) Command Lifecycle Modeling:* This study formally defines the ATCO command lifecycle and introduces a novel framework for jointly predicting Time Offset and Duration, expanding the scope of ATC behavior modeling.

*2) Multimodal Deep Learning:* A CNN-Transformer-based framework is developed to fuse structured data, trajectories, and airspace images, enabling comprehensive multimodal learning.

*3) Interpretability and Application:* The framework incorporates attention-based interpretability, supports real-world deployment, and offers actionable insights for ATCO workload management and decision-making. The code is publicly available.

## II. METHODOLOGY

### A. Dataset

To support ATCO command lifecycle modeling, a multi-source dataset was built by integrating flight trajectories, transcribed voice commands, and contextual information from open-access platforms. All data were aligned by callsigns and timestamps to ensure semantic and temporal consistency.

*1) Trajectory Event Detection*

The trajectory data were collected from a global open ADS-B archive, filtered by geographical bounds and date ranges. Each aircraft's 4D trajectory was represented as a time-series sequence of latitude, longitude, altitude, ground speed, and heading. To associate ATC commands with actual aircraft responses, it was necessary to identify the true initiation times of flight maneuvers from the trajectory data. Based on behavioral patterns, each trajectory was segmented into two phases: stable platforms and change periods. During stable platforms, the aircraft maintained consistent flight parameters and was presumed not to be responding to new commands; during change periods, it actively adjusted its state in response to a command, such as altering altitude, speed, or heading. These change points served as candidate maneuver initiation times and were crucial for command alignment. To extract these maneuver events, a sliding window with histogram-based platform detection method was applied across altitude, speed, and heading data. For each flight parameter, a multi-stage filtering pipeline combined noise suppression (e.g., Savitzky-Golay filtering), adaptive smoothing, and histogram density estimation within a sliding window to identify platform segments. A maneuver onset was defined as the end of a platform where a significant transition to a new state began.

*2) Voice Command Processing*

ATCO voice commands were sourced from a publicly available, manually transcribed dataset containing speaker labels, onset times, and durations. To structure the raw text, natural language processing techniques were applied to extract callsigns, command types, and parameters. Callsigns were identified using regular expressions and normalized via a lookup table mapping airline aliases to ICAO codes. For example, "speedbird one two three turn left heading zero" was parsed and mapped to the standardized callsign BAW123. Commands were categorized into three types: altitude (e.g., "descend to 3000"), speed (e.g., "reduce speed to 210"), and heading (e.g., "turn left heading 180"), covering the majority of tactical instructions in terminal operations. Numerical values were parsed using rule-based methods and encoded as structured integer features. Each command was then represented by a combination of categorical and numerical attributes. To ensure clarity and consistency, compound or conditional commands were excluded. The resulting dataset retained clean, direct control instructions suitable for supervised modeling of maneuver timing.

*3) Feature Engineering*

To capture operational context, auxiliary features were extracted from open datasets, including time-aligned weather (e.g., wind, visibility), airspace structure (e.g., STARs), waypoint density, and historical traffic flow. Aircraft were categorized by wake turbulence class (WTC), and each trajectory point was enriched with spatial cues such as distance/bearing to the airport, route adherence, nearest waypoint, and local traffic density. To enable visual encoding in a multimodal framework, two types of images were generated. Sample images of generated historical trajectories are shown in Fig. 3. For each command timestamp, a 2-minute segment of prior flight path was plotted as a blue line on a standardized coordinate-free image. This allowed the model to consistently interpret heading changes, speed trends, and spatial context. Fig. 4 presents sample images of generated airspace snapshots at the time of command issuance. Each active aircraft is shown as a velocity vector, with the target aircraft highlighted in red and others in blue. This representation conveys local traffic complexity, directional conflicts, and airspace congestion that influence ATCO decision-making.

Finally, Each voice command was aligned to the nearest maneuver in the trajectory based on callsign and timing, capturing both command-to-response delay and maneuver duration. This produced a multimodal dataset with accurately labeled intervals for lifecycle prediction.

### B. LightGBM Baseline Model

As a preliminary experiment, a LightGBM-based regression model was constructed to assess the predictability of the time offset between aircraft maneuvers and ATCO-issued commands.

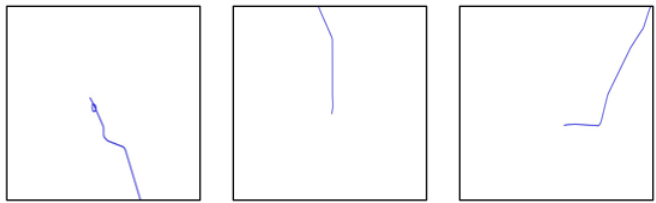

Fig. 3. Sample images of generated historical trajectories.

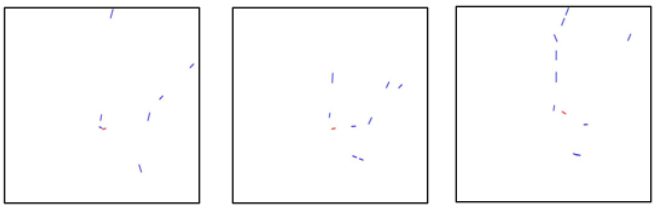

Fig. 4. Sample images of generated airspace snapshots.

This interpretable model was designed to evaluate whether meaningful patterns exist in the data, thereby justifying further deep learning efforts. A set of structured features was used to fit the time offset and generate feature importance rankings. Results showed that the LightGBM model consistently outperformed a naive mean-based baseline, demonstrating that the time offset is not random but statistically predictable, thus validating the modeling objective (see Appendix).

### C. CNN-Transformer Model

A multimodal neural network was developed to jointly model structured variables, historical trajectory sequences, and image-based airspace states, aiming to predict two key variables within the ATCO command lifecycle: time offset and duration. As illustrated in Fig. 5, the model consists of four feature encoding branches and a fusion regression head.

*1) Structured Feature Encoder (MLP)*

The structured feature encoder (MLP_N1) processes categorical and numerical inputs such as flight plans, aircraft models, command parameters, airspace traffic levels, and weather conditions. It consists of two fully connected layers, each followed by Layer Normalization, ReLU activation, and Dropout, yielding a 128-dimensional feature vector.

*2) Spatial Image Encoder (EfficientNet)*

The image feature encoder utilizes EfficientNet-B0 [10] to extract spatial representations from two types of visual inputs: the aircraft's historical trajectory image and the current airspace configuration snapshot. These images are constructed to reflect both localized motion patterns and broader traffic context. EfficientNet-B0, chosen for its balance between accuracy and computational efficiency, employs mobile inverted bottleneck convolution (MBConv) as its core building block. This architecture enables deep feature extraction while maintaining lightweight model complexity, making it well-suited for real-time inference scenarios. In this study, the classification head of EfficientNet-B0 is removed, and the convolutional backbone is retained, including all MBConv layers and pooling operations. Each image is independently processed to produce a 512-dimensional feature vector, capturing spatial complexity relevant to ATCO decision-making.

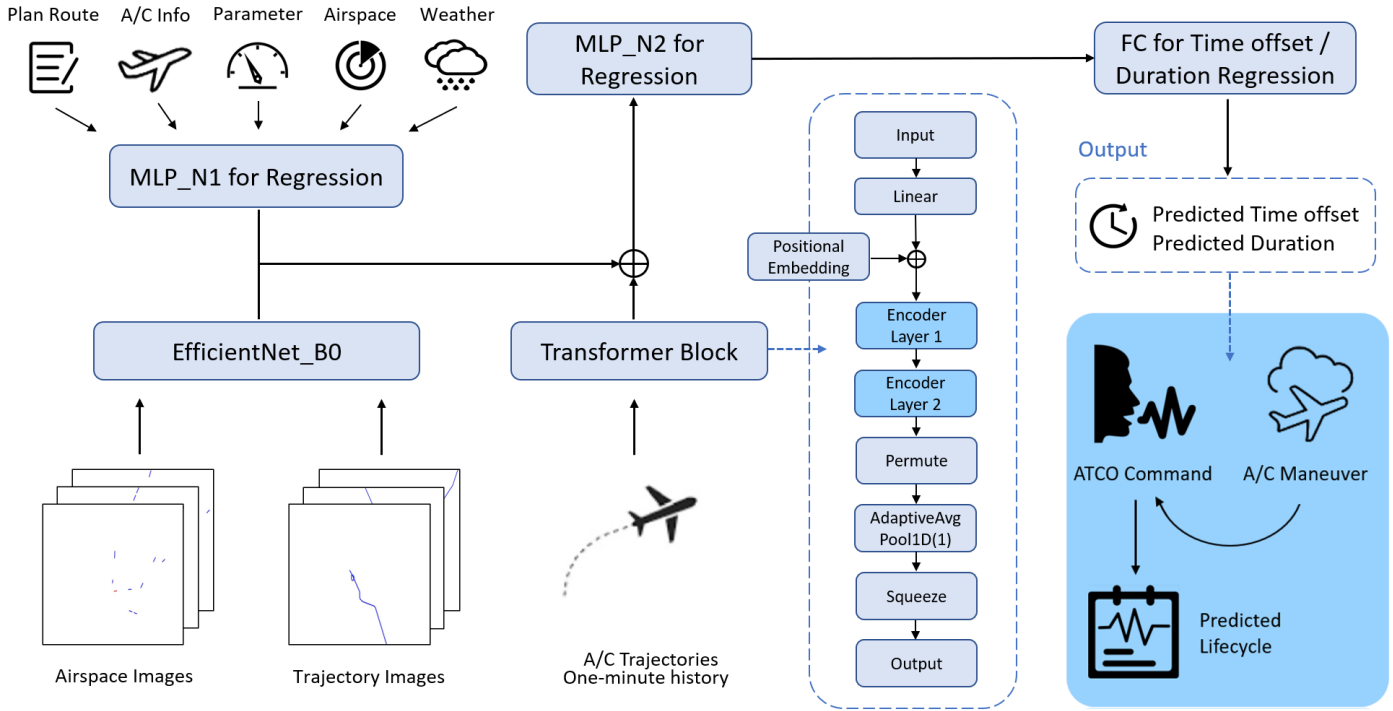

Fig. 5. Architecture of CNN-Transformer model.

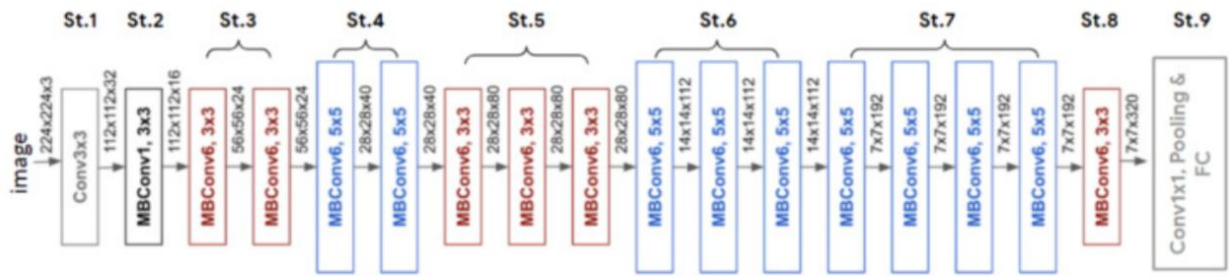

Fig. 6. Architecture of EfficientNet-B0 [10].

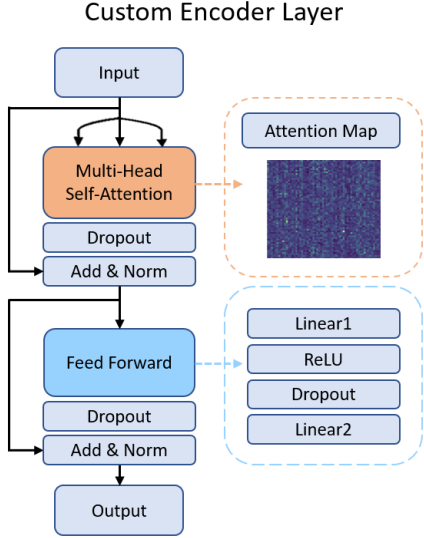

Fig. 7. Architecture of customized encoder layer.

*3) Trajectory Sequence Encoder (Transformer)*

The trajectory sequence encoder captures temporal patterns in the recent flight history, using input sequences of aircraft states such as speed, altitude, and heading from the past 60 seconds. Each sequence is projected into a 128-dimensional hidden space and combined with learnable positional embeddings to retain time order. The encoded sequence is then processed by two custom Transformer encoder layers, shown in Fig. 7. Each layer includes a multi-head self-attention module, residual connections, Layer Normalization, and a feedforward block with a Linear–ReLU–Dropout–Linear structure for non-linear transformation and regularization. Attention maps from the self-attention modules are retained for interpretability. These maps help visualize how the model focuses on different time steps, offering insight into which parts of the flight history influence the predicted command timing. The output is reduced to a 128-dimensional temporal feature vector using Adaptive Average Pooling and a Squeeze operation.

*4) Fusion and Regression Head*

The four encoded feature vectors—128 for structured data, 128 for temporal sequences, 512 for the trajectory image, and 512 for the airspace image—are concatenated into a 1280-dimensional multimodal representation. This is processed by a fusion MLP (MLP_N2), followed by a fully connected layer that outputs the two regression targets: time offset and duration. This setup captures the dynamic link between command issuance and aircraft response, enabling the inverse reconstruction of controller behavior from maneuver events and supporting deeper workload analysis. Cross-modal fusion was also tested using a cross-attention mechanism, where the MLP output acted as the query and the remaining modalities as key and value inputs. However, this method performed slightly worse than simple concatenation. An alternative multi-task learning design, using separate regression heads and loss functions for each target, was also explored but produced less stable results. The final model adopts a joint regression strategy, which better captures the correlation between time offset and duration and offers improved robustness and interpretability.

## III. EXPERIMENTS AND RESULTS

### A. Experiment Settings

All experiments were conducted using PyTorch. The dataset was split into 80% for training and 20% for validation. Structured and sequential features were standardized, and image inputs were augmented using random brightness, contrast adjustment, and Gaussian noise. The model was trained for 200 epochs using the Adam optimizer with a learning rate of 1e-5 and a batch size of 16. CosineAnnealingLR was used for dynamic learning rate adjustment. SmoothL1Loss was used as the loss function for time offset prediction to handle outliers effectively, while MSELoss was chosen for duration due to its sensitivity to fine-grained differences.

To enhance robustness and generalization, an ensemble strategy was adopted. Models were trained under different random seeds and loss configurations. For each regression target—time offset, duration, and overall error—the top two checkpoints based on validation performance were selected from each configuration, resulting in 12 representative models. Final predictions were obtained through weighted averaging. For time offset, higher weights were assigned to models optimized for offset (0.5), followed by overall (0.3) and duration (0.1); a mirrored scheme was used for duration. Fig. 8 illustrates the ensemble strategy based on weighted averaging. This task-aware weighting improved performance across both targets while reducing sensitivity to data variation.

### B. Evaluation Metrics

Model performance was evaluated using multiple regression metrics, covering both overall and variable-specific performance. The primary metrics include Mean Absolute Error (MAE), Root Mean Square Error (RMSE), and the coefficient of determination ($R^2$). Definitions are as follows:

$$MAE_{overall} = \frac{1}{2N}\sum_{i=1}^{N}\left(\left|y_i^{(1)} - \widehat{y_i^{(1)}}\right| + \left|y_i^{(2)} - \widehat{y_i^{(2)}}\right|\right) \quad (1)$$

$$RMSE_{overall} = \sqrt{\frac{1}{2N}\sum_{i=1}^{N}\left[\left(y_i^{(1)} - \widehat{y_i^{(1)}}\right)^2 + \left(y_i^{(2)} - \widehat{y_i^{(2)}}\right)^2\right]} \quad (2)$$

$$R^2_{overall} = 1 - \frac{\sum_{i=1}^{N}\left[\left(y_i^{(1)} - \widehat{y_i^{(1)}}\right)^2 + \left(y_i^{(2)} - \widehat{y_i^{(2)}}\right)^2\right]}{\sum_{i=1}^{N}\left[\left(y_i^{(1)} - \overline{y^{(1)}}\right)^2 + \left(y_i^{(2)} - \overline{y^{(2)}}\right)^2\right]} \quad (3)$$

To evaluate performance at the variable level, all metrics were further decomposed into sub-metrics for time offset and duration individually, including MAE, RMSE, and $R^2$ for both output variables.

### C. Comparative Experiment

To evaluate overall performance, the ensemble model was compared with three baselines: LightGBM, TabPFN [11], and the best single model by validation score. TabPFN, a transformer-based model for tabular data, is known for fast generalization without fine-tuning. As shown in Table I, the ensemble achieved the highest overall R² score (0.19), indicating better generalization and fit. While the best single model slightly outperformed in MAE, the ensemble yielded much higher R² scores for time offset (0.27) and duration (0.11), suggesting better trend capture and less overfitting. In contrast, TabPFN performed poorly across all metrics, with negative R² values. The ensemble was selected as the final predictor for its balance of accuracy, stability, and interpretability, suitable for both analysis and deployment.

### D. Ablation Study

Ablation experiments were conducted to assess the impact of each input modality and CNN architecture. Table II reports the results. Removing any branch led to increased error, confirming the benefit of multimodal fusion. Airspace and trajectory images improved offset prediction, while the Transformer was more effective for duration. In terms of CNN architecture, EfficientNet-B0 outperformed both ResNet18 [12] and a custom shallow CNN, highlighting the importance of deep, efficient visual encoding in capturing spatial complexity.

TABLE I. MODEL PERFORMANCE COMPARISON

| Model | MAE_overall | RMSE_overall | R²_overall | MAE_offset | MAE_duration | RMSE_offset | RMSE_duration | R²_offset | R²_duration |
|---|---|---|---|---|---|---|---|---|---|
| LightGBM | 4.71 | 8.40 | 0.16 | 8.45 | 0.98 | 11.80 | 1.41 | 0.16 | 0.16 |
| TabPFN | 6.46 | 11.08 | -0.57 | 11.51 | 1.42 | 15.54 | 1.99 | -0.46 | -0.69 |
| Best Single (Ours) | 4.63 | 7.70 | 0.16 | 8.24 | 1.02 | 10.76 | 1.64 | 0.26 | 0.05 |
| **Ensemble (Ours)** | 4.89 | 7.73 | **0.19** | 8.67 | 1.10 | 10.81 | 1.61 | **0.27** | **0.11** |

TABLE II. ABLATION STUDY RESULT

| CNN Type | MLP | Transformer Block | Trajectory Images | Airspace Images | MAE_overall | MAE_offset | MAE_duration |
|---|---|---|---|---|---|---|---|
| efficientnet | ✓ | ✓ | ✓ | ✓ | **4.86** | **8.48** | **1.24** |
| efficientnet | ✓ | ✓ | ✓ | ✗ | 5.21 | 9.08 | 1.35 |
| efficientnet | ✓ | ✓ | ✗ | ✓ | 5.22 | 9.24 | 1.20 |
| efficientnet | ✓ | ✗ | ✓ | ✓ | 4.98 | 8.42 | 1.53 |
| efficientnet | ✗ | ✓ | ✓ | ✓ | 4.98 | 8.61 | 1.35 |
| efficientnet | ✓ | ✓ | ✗ | ✗ | 7.20 | 12.63 | 1.76 |
| resnet | ✓ | ✓ | ✓ | ✓ | 5.56 | 9.76 | 1.37 |
| resnet | ✓ | ✓ | ✓ | ✗ | 5.57 | 9.89 | 1.24 |
| resnet | ✓ | ✓ | ✗ | ✓ | 5.36 | 9.47 | 1.25 |
| resnet | ✓ | ✗ | ✓ | ✓ | 5.58 | 9.94 | 1.22 |
| resnet | ✗ | ✓ | ✓ | ✓ | 5.47 | 9.74 | 1.20 |
| resnet | ✓ | ✓ | ✗ | ✗ | 7.36 | 12.98 | 1.74 |
| custom | ✓ | ✓ | ✓ | ✓ | 5.84 | 10.50 | 1.18 |
| custom | ✓ | ✓ | ✓ | ✗ | 6.21 | 11.12 | 1.30 |
| custom | ✓ | ✓ | ✗ | ✓ | 5.96 | 10.71 | 1.21 |
| custom | ✓ | ✗ | ✓ | ✓ | 5.95 | 10.56 | 1.33 |
| custom | ✗ | ✓ | ✓ | ✓ | 5.93 | 10.49 | 1.38 |
| custom | ✓ | ✓ | ✗ | ✗ | 7.23 | 12.36 | 2.09 |

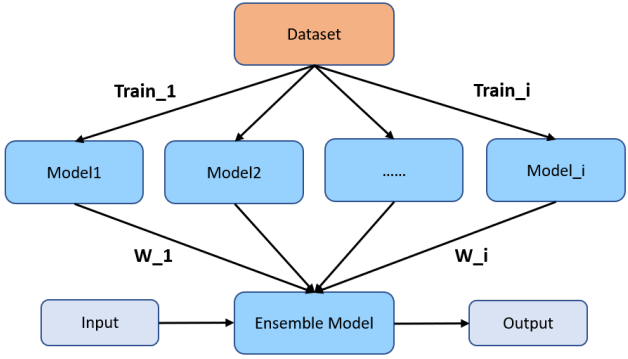

Fig. 8. Weighted model ensemble strategy.

*E. Interpretability Analysis*

To better understand how each input modality contributes to the model's predictions, a series of interpretability analyses were conducted using SHAP and attention-based visualization techniques.

For structured inputs, SHAP analysis revealed that features related to command type and aircraft motion had the highest influence on model predictions. In the case of time offset, the most important feature was whether the command was speed-related (velocity), with an average SHAP value of +0.09. Other key contributors included heading commands (head), calibrated airspeed (cas), and the aircraft's bearing to the airport (bearing_to_airport), reflecting the model's sensitivity to aircraft dynamics and spatial positioning. In contrast, contextual variables such as weather conditions and peak-hour indicators had negligible contributions, suggesting they offered little discriminative value. These findings are consistent with the LightGBM baseline analysis. For duration, velocity remained the most relevant variable, though its importance decreased significantly (average SHAP +0.03), indicating that speech duration is less strongly tied to structured inputs and may depend more on latent factors such as phrase structure or controller habits. Other modestly contributing features included distance to the airport and flight level, hinting at a tendency for controllers to use longer phrases when communicating with distant or high-altitude aircraft. Most other features showed low overall impact, though some—such as traffic density or planned routing—had occasional localized effects under specific conditions. Beeswarm plots further highlighted the directional impact of each feature. For instance, velocity exhibited a clear binary pattern in time offset prediction, with speed commands increasing predicted delay, while for duration, it had an opposite effect. These results confirm that the model's attention to structured features aligns with operational intuition and provide a basis for future pruning of low-impact variables to improve model efficiency.

For temporal inputs, attention maps from the Transformer encoder revealed how the model distributes focus across time steps. In earlier layers, attention was dispersed, while deeper layers selectively emphasized key moments in the trajectory, such as turning points or speed changes. This progression confirms the Transformer's capacity to model temporal dependencies and identify behaviorally significant events.

For image inputs, Grad-CAM was applied to visualize activation regions in both current airspace snapshots and historical trajectory images. The model consistently attended to areas with high traffic density or recent maneuvers, indicating that the CNN modules effectively capture spatial cues that enhance the model's understanding of airspace complexity.

Overall, these interpretability results underscore the complementary contributions of all three modalities and demonstrate that the model's predictions are based on semantically meaningful patterns. They also provide further evidence supporting the effectiveness of multimodal fusion. Full SHAP plots, attention heatmaps, and Grad-CAM visualizations are provided in the Appendix.

*F. Case Study*

To further illustrate the model's prediction capability, two representative case studies on ATCO command lifecycle prediction are presented, covering both single-command and high-density multi-command scenarios.

*1) Single Command Prediction*

An example involving flight QFA1 is first analyzed. Fig. 9 shows speed variation over time for this representative flight. At approximately 50,789 seconds, an ATCO issued a speed reduction command from 250 knots to 220 knots. Around 50,811 seconds, the aircraft executed a corresponding deceleration maneuver, which the model successfully identified. A timeline-based visualization of this lifecycle is presented in Fig. 10, showing the actual voice segment (blue), predicted voice duration (yellow), and observed maneuver (green). The model's predictions closely align with the true sequence, with a duration prediction error of only 0.1 seconds in this instance. This result demonstrates the model's ability to capture the temporal structure of ATCO behavior with high fidelity.

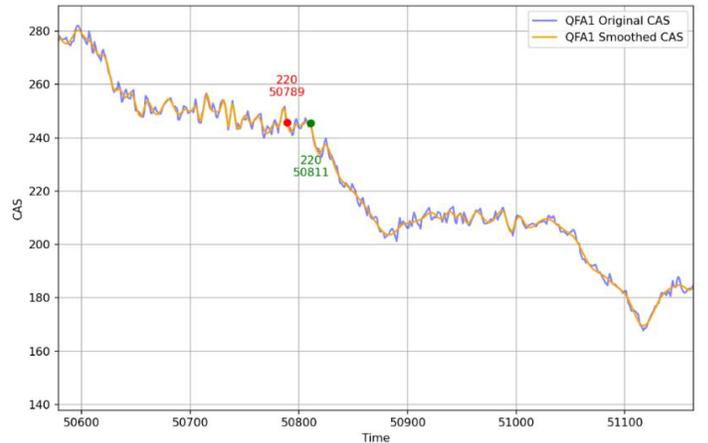

Fig. 9. Command and actual maneuver timestamps for a flight during a speed change event. The red marker indicates the ATCO command time, and the green marker indicates the observed maneuver time.

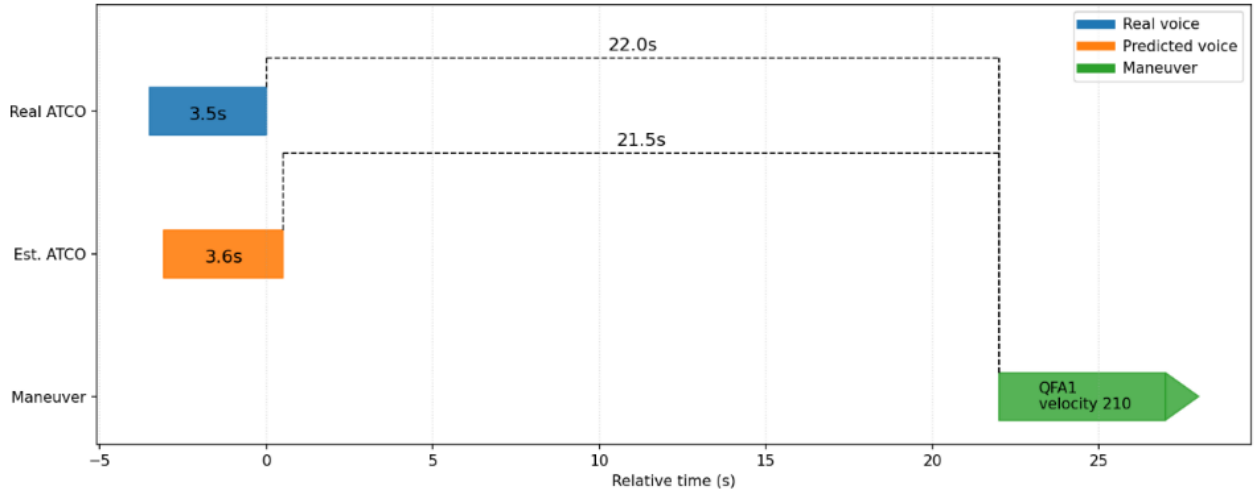

Fig. 10. Visualization of the predicted command lifecycle. The blue bar represents the actual voice command duration from the controller, while the orange bar shows the predicted duration. The green arrow indicates the aircraft maneuver (e.g., QFA1 adjusting velocity to 210 knots). Dashed lines illustrate the time offset between the end of the voice command and the start of the maneuver, based on both real and estimated values.

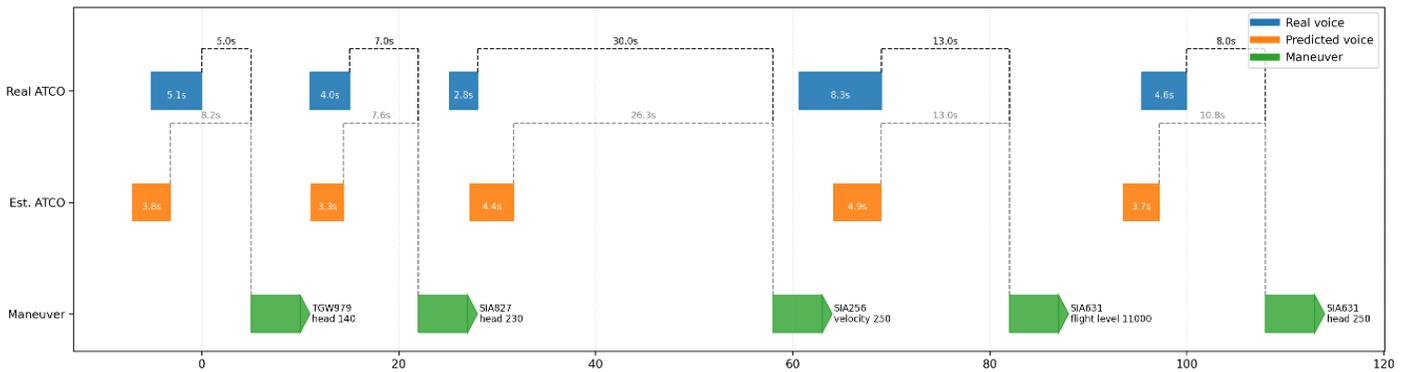

Fig. 11. Visualization of the predicted command lifecycle in a high-load airspace window.

*2) Multi-Command Prediction*

To evaluate model generalization in dense command scenarios, a high-load terminal area window spanning 100 seconds was selected. Fig. 11 presents the model's predicted command lifecycles for several flights during this interval, including TGW979 (heading 140), SIA827 (heading 230), SIA256 (velocity 250), and SIA631 (flight level 11,000; heading 250). Despite the close temporal proximity of commands, the model was able to reconstruct each lifecycle with accurate alignment between predicted voice timing and maneuver onset. These examples confirm the model's robustness in handling high-density, multi-target scenarios and its potential for supporting real-time workload analysis.

## IV. LIMITATIONS

Despite the effectiveness of the proposed CNN-Transformer ensemble model in predicting ATCO command lifecycles, several limitations remain:

*1) CDO Handling:* The model cannot accurately predict command timing during Continuous Descent Operations (CDO), where flights often lack distinct level-off segments. As a result, it is difficult to identify clear maneuver points. To maintain modeling accuracy, CDO phases were excluded from this study. Future work should incorporate fine-grained vertical trend analysis and descent rate modeling to improve support for CDO trajectories.

*2) Conditional Commands:* Some controller instructions include execution conditions (e.g., "descend after passing waypoint X"), which introduce delayed maneuvers. These commands disrupt the direct mapping between voice and trajectory, leading to large time offset variance or label noise. Although such cases were excluded from the current dataset, future models should integrate speech content analysis and trajectory event alignment to support conditional execution logic.

*3) Single-Command Assumption:* The model assumes that each voice segment corresponds to a single command. However, ATCOs frequently issue multiple instructions in a single transmission. Using one-hot encoding for command type limits the model's expressiveness. Future directions include multi-label command encoding and semantic segmentation of composite voice inputs.

*4) Misalignment Caused by Overlapping Commands:* In some cases, a new command is issued before the aircraft completes the previous maneuver. This may cause the aircraft to bypass intermediate phases, such as level flight, resulting in missing or misaligned lifecycle targets. Future work could introduce command queue modeling and transition-state reasoning to resolve such discontinuities.

## V. Conclusion

This study proposes a multimodal deep learning framework to model the lifecycle of air traffic control (ATCO) commands and estimate controller workload in terminal maneuvering areas using aircraft 4D trajectories. Trajectory data were preprocessed through filtering, and maneuver points were detected using a sliding window with histogram-based methods. A LightGBM model first validated task feasibility and identified key features. Building on this, a CNN-Transformer model was developed to predict two temporal variables: the time offset between command issuance and maneuver execution, and command duration. The model integrates structured flight and environmental data, trajectory sequences, and airspace representations rendered as images. Attention map visualizations enhance interpretability. Comparative and ablation experiments demonstrated the independent and complementary value of each modality. Image and trajectory inputs were shown to play distinct yet synergistic roles in reconstructing the timing of ATCO decisions. The model can infer command lifecycles from flight behavior alone, enabling timeline reconstruction of ATCO activity without relying on raw audio data.

This work lays a foundation for automated command generation and workload estimation, with applications in ATCO resource planning, airspace management, and flight scheduling. Future extensions may support real-time controller assistance in collaborative decision-making systems. Due to space limits, only a condensed version is presented; further results and visualizations are available upon request.

## References

[1] S. G. Hart, "NASA-task load index (NASA-TLX); 20 years later," in *Proceedings of the Human Factors and Ergonomics Society Annual Meeting*, vol. 50, pp. 904–908, 2006.

[2] I. V. Laudeman, S. G. Shelden, R. Branstrom, and C. L. Brasil, *Dynamic Density: An Air Traffic Management Metric*, NASA Tech. Memo. 112226, Apr. 1998.

[3] G. Tobaruela, W. Schuster, A. Majumdar, W. Y. Ochieng, L. Martinez, and P. Hendrickx, "A method to estimate air traffic controller mental workload based on traffic clearances," *Journal of Air Transport Management*, vol. 39, pp. 59–71, 2014.

[4] D. Gianazza, "Learning air traffic controller workload from past sector operations," in Proc. 12th USA/Europe Air Traffic Management R&D Seminar (ATM Seminar), Seattle, WA, USA, Jun. 2017.

[5] G. G. Teuler, R. M. Arnaldo, V. F. Gómez, P. M. López, and R. R. Rodríguez, "Study of the impact of traffic flows on the ATC actions," *Aerospace*, vol. 9, no. 8, Art. no. 467, 2022.

[6] H. Xie, M. Zhang, J. Ge, X. Dong, and H. Chen, "Learning air traffic as images: A deep convolutional neural network for airspace operation complexity evaluation," *Complexity*, vol. 2021, Art. no. 6457246, 2021.

[7] Y. Pang, J. Hu, C. S. Lieber, N. J. Cooke, and Y. Liu, "Air traffic controller workload level prediction using conformalized dynamical graph learning," *Advanced Engineering Informatics*, vol. 57, Art. no. 102113, 2023.

[8] M. C. Shyr, A. H. Farrahi, and S. Verma, "Predictive workload model for air traffic controllers during UAM operations," in *Proc. AIAA/IEEE 43rd Digit. Avion. Syst. Conf. (DASC)*, San Diego, CA, USA, 2024, pp. 1–6.

[9] J. Yang, H. Yang, Z. Wu, and X. Wu, "Cognitive load assessment of air traffic controller based on SCNN-TransE network using speech data," *Aerospace*, vol. 10, no. 7, Art. no. 584, 2023.

[10] M. Tan and Q. Le, "EfficientNet: Rethinking model scaling for convolutional neural networks," in Proc. 36th Int. Conf. Mach. Learn. (ICML), Proc. Mach. Learn. Res., vol. 97, pp. 6105–6114, 2019.

[11] N. Hollmann, S. Müller, L. Purucker, A. Krishnakumar, M. Körfer, S. B. Hoo, *et al.*, "Accurate predictions on small data with a tabular foundation model," *Nature*, vol. 637, no. 8045, pp. 319–326, 2025.

[12] K. He, X. Zhang, S. Ren, and J. Sun, "Deep residual learning for image recognition," in Proc. IEEE Conf. Comput. Vis. Pattern Recognit. (CVPR), Las Vegas, NV, USA, 2016, pp. 770–778.

## Appendix

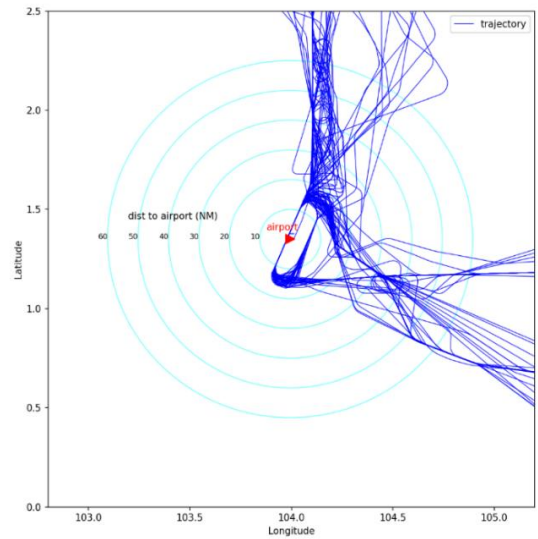

Fig. A.1. Trajectories of arriving aircraft within one day at an airport

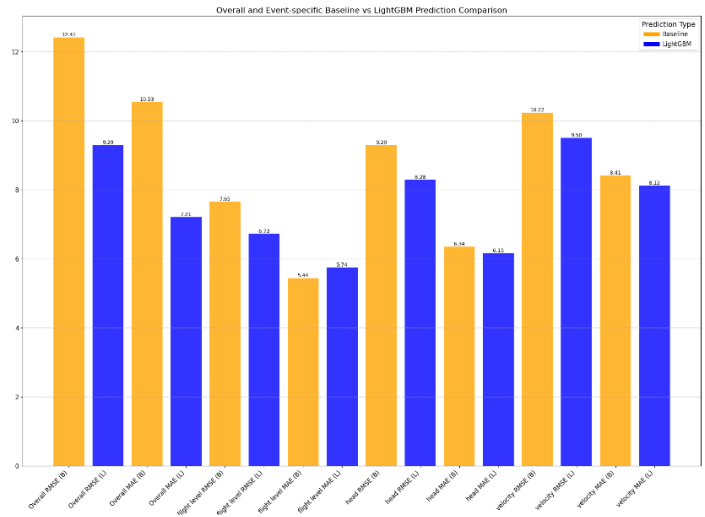

Fig. A.2. LightGBM Performance Comparison. The yellow bar indicates the mean-based baseline; the blue bar represents LightGBM.

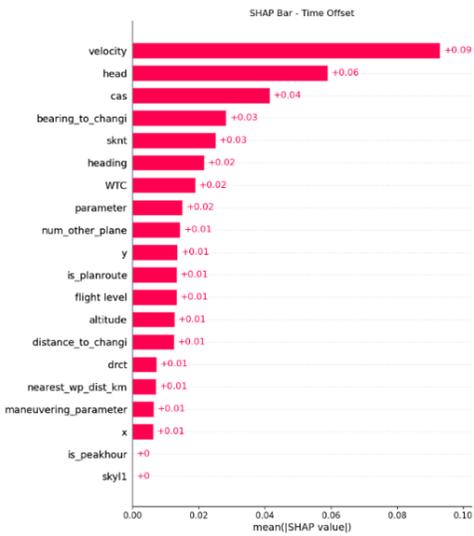

Fig. A.3. SHAP feature importance bar chart — Time Offset.

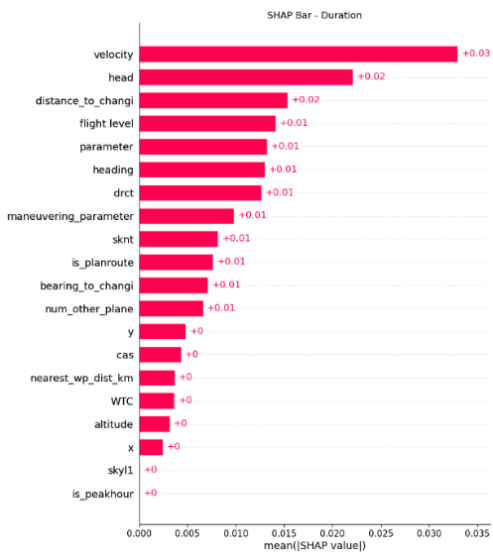

Fig. A.4. SHAP feature importance bar chart — Duration.

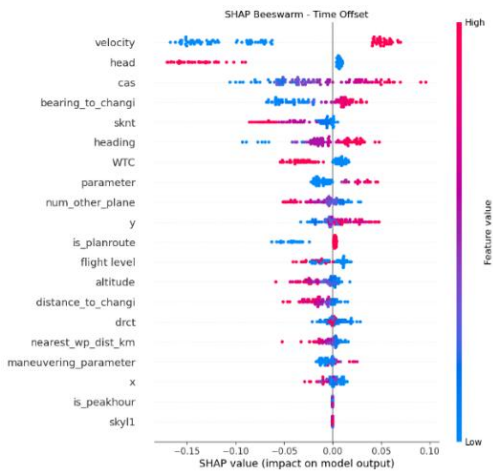

Fig. A.5. SHAP beeswarm plot — Time Offset.

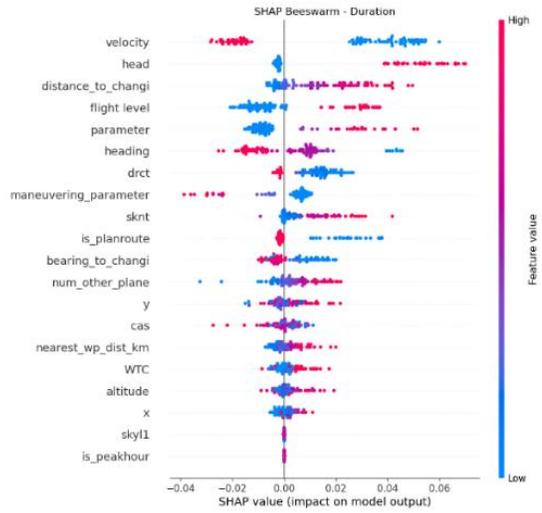

Fig. A.6. SHAP beeswarm plot — Duration.

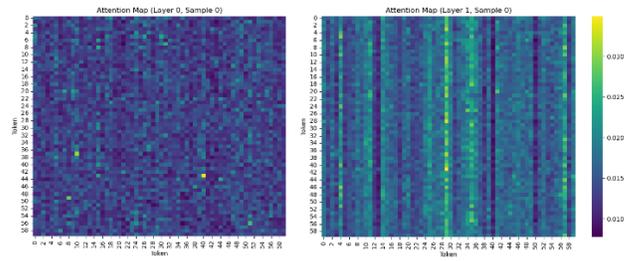

Fig. A.7. Attention heatmaps of the first and second layers in the customized Transformer module.

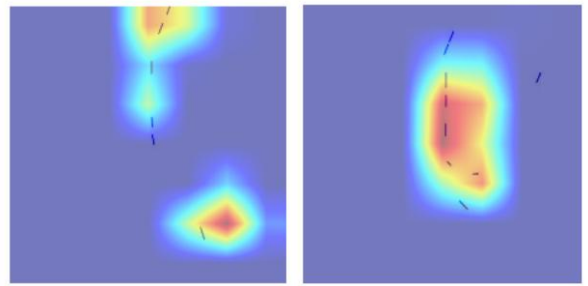

Fig. A.8. Grad-CAM heatmap of the airspace snapshots.

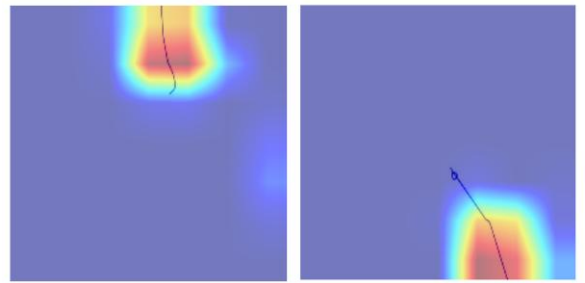

Fig. A.9. Grad-CAM heatmap of the historical trajectory images.